\pdfoutput=1

\documentclass{article}

\usepackage{corl_2026}

\usepackage[utf8]{inputenc}
\usepackage[T1]{fontenc}
\usepackage{url}
\usepackage{booktabs}
\usepackage{amsfonts}
\usepackage{nicefrac}
\usepackage{microtype}
\usepackage{xcolor}
\usepackage{graphicx}
\usepackage{amsmath,amssymb}

\usepackage{tikz}
\usetikzlibrary{scopes,patterns,arrows.meta,positioning,fit,calc,backgrounds}
\usepackage{float}
\usepackage{tabularx}
\newcolumntype{L}{>{\raggedright\arraybackslash}X}

\hypersetup{%
  pdftitle={KlaskTron-IL: A Contact-Rich Adversarial Imitation Learning Benchmark with a Validated Sim-to-Real Bridge},
  pdfauthor={Nicklas Sohm, Aswin Ramachandran, Raffaello D'Andrea}%
}

\title{KlaskTron-IL: A Contact-Rich Adversarial Imitation Learning Benchmark with a Validated Sim-to-Real Bridge}

\author{
  Nicklas Sohm\\
  Aswin Ramachandran\\
  Raffaello D'Andrea\\
  ETH Zürich, IDSC\\
  \texttt{aramachandra@ethz.ch}
}

\begin{document}
\maketitle

\begin{abstract}
KlaskTron is an open-hardware physical platform for two-player adversarial reinforcement learning, with a learned actuator model that supports zero-shot sim-to-real transfer of trained policies. Building on this platform, we contribute KlaskTron-IL: a multi-tier imitation learning dataset (80k human demonstration timesteps collected directly on the physical robot, 320k symmetry-augmented timesteps, and synthetic expert trajectories up to 5M steps generated by the zero-shot-validated PPO policy), and diagnostic baselines that expose three characteristic failure modes of standard IL methods under contact-rich adversarial dynamics: BC's data scaling plateau, GAIL's interpretable but weak-gradient discriminator (Spearman $\rho = -1.0$ peg--ball proximity recovery, but a logit span under 0.1), and an asymmetric BC-prior $\times$ reward-density interaction. Because the policy interface used throughout this paper is the same one the actuator model bridges to hardware, a policy trained on this benchmark can run unchanged on the physical robot, as we demonstrate zero-shot for the strongest BC baseline, so long as its commands respect the actuator model's $\pm0.2$\,m/s identification range.
\end{abstract}

\keywords{Imitation Learning, Sim-to-Real, Adversarial Multi-Agent RL, Benchmark}

\section{Introduction}
\label{sec:intro}

Physical adversarial multi-agent learning is bottlenecked by infrastructure. Industrial-grade platforms such as robotic table tennis~\citep{table_tennis} are inaccessible to most labs; cooperative manipulation benchmarks~\citep{pmlr-v164-mandlekar22a,pmlr-v229-walke23a,jia2024towards} avoid the high-speed contact and non-stationarity that adversarial play induces. KlaskTron addresses this with a $<\$2{,}500$ open-hardware testbed based on the tabletop game Klask~\citep{Klask}: dual CoreXY gantries drive magnetic pegs beneath the board at 50\,Hz, an overhead camera with EKF state estimation closes the perception loop, and a learned neural actuator model~\citep{hwangbo} bridges the simulator and the physical robot. Crucially, the platform supports \emph{zero-shot} sim-to-real transfer: PPO policies trained in Isaac Lab deploy directly to hardware without further fine-tuning, even with domain randomization disabled~\citep{VenkatapathyA-RSS-26}.

This paper asks what happens when standard imitation learning is run on this platform. The same actuator network that enables PPO transfer also makes a policy with the same interface (12D state, 2D-velocity action) hardware-deployable, provided its commands stay within the action range the actuator model was identified on, so failure modes diagnosed in simulation are directly informative about what would fail on the robot. Three things go wrong, and they go wrong in interpretable ways: (a) Behavioral cloning exhibits a \textbf{data-scaling plateau} across 80k to 5M expert demonstrations, with performance statistically flat and well below expert parity regardless of dataset size -- a signature that covariate shift, not data scarcity, is the binding constraint; (b) GAIL learns an \textbf{interpretable but weak-gradient discriminator} that recovers peg--ball proximity with perfect rank correlation (Spearman $\rho = -1.0$) but whose logit span is too narrow to serve as an informative reward in our configuration; (c) BC-bootstrapped RL displays an \textbf{asymmetric demonstration-prior $\times$ reward-density interaction} -- the BC prior stabilizes sparse-reward training but has no detectable effect under dense shaping.

Our contributions are:
\begin{enumerate}
    \item A multi-tier IL dataset: 80k human demonstration timesteps recorded directly on the physical KlaskTron robot via EKF-based state estimation, 320k symmetry-augmented timesteps, and synthetic expert trajectories at four scales (80k/250k/1M/5M) generated by a PPO policy trained under the curriculum of~\citep{VenkatapathyA-RSS-26} and itself deployed zero-shot on hardware (Section~\ref{sec:hardware}).
    \item Diagnostic baselines (BC, GAIL, BC+RL) that fail informatively, exposing the three named failure modes summarized in Table~\ref{tab:baselines_summary} and detailed in Section~\ref{sec:baselines}.
\end{enumerate}

Because the dataset is collected on the physical robot, and because the synthetic data is generated by a policy with a verified path to hardware, KlaskTron-IL is the only adversarial IL benchmark we know of where algorithmic progress in simulation has a demonstrated path to real-world hardware (Section~\ref{sec:hardware}), provided the method's commands stay within the actuator model's identification range.

\textbf{Dataset:} \url{https://huggingface.co/datasets/KlaskLab/klasktron-il-benchmark} (CC-BY-4.0). \textbf{Code:} \url{https://github.com/IDSCETHZurich/klasktron-il} (Apache-2.0). The repository contains the Isaac Lab environment, training pipelines for all baselines, the evaluation harness, per-experiment configuration files, and the inherited actuator model and PPO expert needed to reproduce the evaluations.

\section{Related Work}
\label{sec:related}

\textbf{IL datasets and offline RL benchmarks.} RoboMimic~\citep{pmlr-v164-mandlekar22a} introduced quality-stratified demonstrations for systematic IL evaluation; BridgeData~V2~\citep{pmlr-v229-walke23a} scaled to over 60k cross-embodiment trajectories; D3IL~\citep{jia2024towards} targeted behavioral diversity; and the IL Datasets toolkit~\citep{gavenski2024ildatasets} standardized HuggingFace dataset creation. Offline RL has produced parallel infrastructure: D4RL~\citep{fu2021d4rldatasetsdeepdatadriven}, RL Unplugged~\citep{RL_Unplugged}, and Minari~\citep{younis_2024_13767625}. None of these environments work against the agent. KlaskTron-IL introduces adversarial dynamics into the IL dataset paradigm: the state distribution a policy visits is shaped by an opponent, not only by its own errors.

\textbf{Adversarial and contact-rich physical benchmarks.} CyberRunner~\citep{sample_efficient} demonstrated that low-cost open hardware can drive dynamic single-agent RL. Robot air hockey~\citep{jankowski2024airlihockeyhighlyreactivecontact}, high-speed table tennis~\citep{table_tennis}, and foosball platforms~\citep{Moos_2024} push contact dynamics and reaction-time limits. KlaskTron is two-player adversarial (unlike single-agent dynamic platforms), open-hardware at under \$2{,}500 (unlike industrial table-tennis stacks), and -- critically for this paper -- ships a curated demonstration dataset that prior adversarial platforms lacked.

\textbf{Covariate shift, adversarial IL, and demonstration-bootstrapped RL.} Ross et al.~\citep{Dagger} showed that BC error compounds quadratically with horizon length, motivating DAgger and its successors~\citep{thriftydagger,lazydagger}. Stable BC~\citep{Stable_BC} adds control-theoretic stability guarantees; distributionally-robust BC~\citep{NEURIPS2024_c556da88} trains against worst-case shifts; diffusion policies~\citep{Chi-RSS-23} model multi-modal action distributions. GAIL~\citep{GAIL} frames imitation as a minimax game; diffusion-based variants~\citep{DRAIL} address its brittleness. Demonstrations have also been used to warm-start RL~\citep{hansen2023modem}. What is missing is empirical evaluation under adversarial dynamics, where an opponent is part of the dynamics. KlaskTron-IL's data-scaling ladder (80k to 5M steps) and 2$\times$2 reward-density structure provide controlled diagnostics for this regime.

\section{The KlaskTron Platform and Simulation Environment}
\label{sec:platform}

\begin{figure}[t]
    \centering
    \resizebox{0.95\linewidth}{!}{\input{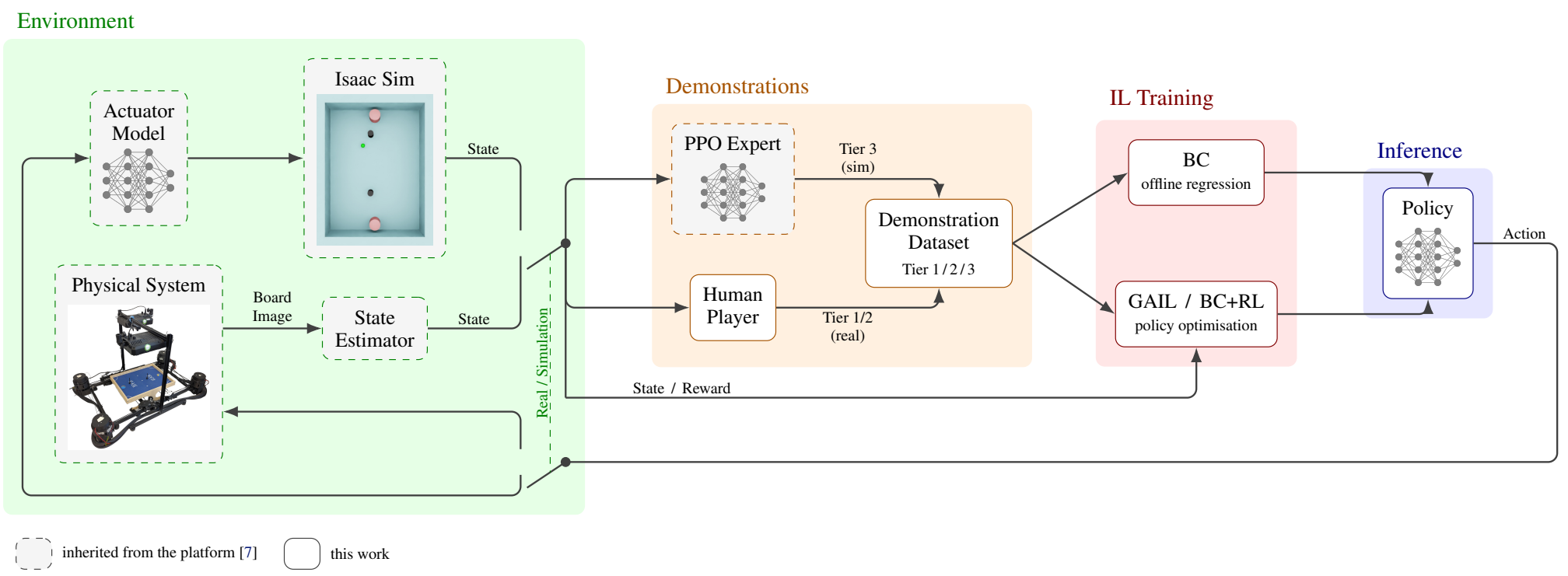}}
    \caption{KlaskTron-IL system architecture. Dashed blocks are inherited from the platform~\citep{VenkatapathyA-RSS-26}. Solid blocks are contributions of this paper. The simulator and physical robot expose the same 12D state / 2D velocity interface, and a learned actuator model lets a single trained policy run unchanged in either. Demonstrations come from two sources: the inherited PPO expert acting in simulation (Tier 3, Section~\ref{sec:dataset}) and human play recorded on the physical robot (Tiers 1--2). BC learns offline from demonstrations alone, GAIL and BC+RL additionally optimize against the environment (Section~\ref{sec:baselines}).}
\label{fig:system_arch}
\end{figure}

We summarize the platform here; full hardware design and sim-to-real validation are described in concurrent work~\citep{VenkatapathyA-RSS-26}. The physical system, the simulator, the learned actuator model, and the PPO agent we use as a demonstration source are inherited from that work unchanged. The demonstration tiers, the imitation-learning baselines, and the evaluation protocol are contributions of this paper (Figure~\ref{fig:system_arch}).

\textbf{Physical platform.} Two CoreXY gantries with direct-drive brushless motors actuate magnetic carriages under a standard Klask board at 50\,Hz. State is estimated from an overhead camera through an EKF with adaptive process noise for collision handling, producing a 12D vector (positions and velocities of both pegs and the ball). The total hardware cost is under \$2{,}500 with off-the-shelf components and 3D-printed parts.

\textbf{Simulation environment.} The Isaac Lab digital twin~\citep{VenkatapathyA-RSS-26} is GPU-native and supports massively parallel rollouts (20M PPO steps in $\sim$12 minutes on a single RTX 4090). The PhysX backend models board geometry, wall collisions, and peg-ball contact with tunable friction and restitution parameters. The observation space is 12-dimensional; each agent outputs a 2D velocity command $a = (v_x, v_y)$ in m/s. The environment imposes no action limit of its own, and each method sets its own bound. The inherited PPO expert is clipped at $\pm 0.2$\,m/s (the bound used by each baseline is given in Section~\ref{sec:baselines}).
In self-play, both agents' commands concatenate into a 4D joint action. Episodes terminate on goals, peg-in-goal events, or a 5\,s timeout (250 steps).

\textbf{Validated zero-shot sim-to-real bridge.} The transfer mechanism is a learned actuator network~\citep{hwangbo}, identified on the physical robot and adopted unchanged from~\citep{VenkatapathyA-RSS-26}, that replaces the idealized PD controller in simulation and captures friction and latency of the brushless motors. The platform paper~\citep{VenkatapathyA-RSS-26} demonstrates that policies trained against this actuator transfer zero-shot to hardware, including with domain randomization disabled. \emph{This is the property that makes KlaskTron-IL's diagnostic findings actionable: any algorithm that improves on the baselines in Section~\ref{sec:baselines}, while respecting the 12D state / 2D velocity interface and the $\pm 0.2$\,m/s command range the actuator model was identified on, inherits the same hardware path.}

For the scoping of this paper, we omit the magnetic biscuit obstacles present in the full Klask game; this isolates contact-rich adversarial dynamics from combinatorial rule complexity, a deliberate choice for a first-pass IL diagnostic. We discuss this and other limitations in Section~\ref{sec:limitations}.

\section{The KlaskTron-IL Dataset}
\label{sec:dataset}

The dataset is organized into three tiers (Table~\ref{tab:dataset_stats}), spanning a spectrum from scarce but authentic human play recorded on the physical robot to abundant synthetic expert data from the zero-shot-validated policy. All tiers expose the identical 12D state / 4D joint-action format, matching the physical-robot interface.

\textbf{Tier~1: Human Demonstrations on Hardware (80k timesteps).} 27 minutes of gameplay (80,591 timesteps across 111 episodes) recorded on the physical KlaskTron robot during a lab tournament with recreational-to-intermediate players. State estimation used the same EKF/AprilTag pipeline that runs at policy inference time, so Tier~1 records exactly what a deployed policy would observe. The 80k step budget represents what is practically collectible in a single physical-robot session.

\emph{Known limitation.} Human actions are reconstructed from finite-differenced position estimates ($a_t \approx \Delta p / \Delta t$), which makes the action labels nearly deterministically predictable from consecutive observations and constitutes a data-leakage risk for any IL method consuming velocity components of the state. The mitigation we use throughout Section~\ref{sec:bc_scaling} is to zero out velocity observations during BC training. Users of Tier 1 should apply the same mitigation or verify their method is invariant to this channel.

\textbf{Tier~2: Symmetry-Augmented Human Demonstrations (320k timesteps).} Exploiting the board's bilateral symmetry, we mirror each Tier~1 trajectory across the $x$ and $y$ axes, yielding 444 episodes and 322,364 timesteps. The augmented trajectories are physically valid because the board geometry is symmetric. Tier~2 isolates the effect of geometric data augmentation on IL robustness.

\textbf{Tier~3: Synthetic Expert from a Zero-Shot-Validated Policy (5M timesteps).} A single file of 5,000,000 timesteps (34,649 episodes, avg.~$\sim$144 steps) generated by the PPO expert in self-play. The scaling study in Section~\ref{sec:bc_scaling} uses prefixes of this file at 80k, 250k, 1M, and 5M steps. Importantly, the generating policy \emph{runs zero-shot on the physical robot} (Section~\ref{sec:hardware}). This means Tier~3 is not merely "synthetic" in the conventional simulation-only sense, but a synthetic trace of a hardware-deployable behavior. The agent reaches an Elo of 1283 (61.8\% win rate vs.\ a 1200 baseline) and wins 10\% of games against a recreational human; it has consistent reaction times but exhibits 0.5--1.5\,cm offensive targeting errors, making its action distribution meaningfully different from human play.

\begin{table}[t]
\caption{KlaskTron-IL dataset tiers. All tiers share the 12D state / 4D joint-action format used on the physical robot and are released under CC-BY-4.0. The synthetic tier ships as a single 5M-step file; the scaling study (Section~\ref{sec:bc_scaling}) uses prefixes.}
\label{tab:dataset_stats}
\centering
\small
\begin{tabular*}{\textwidth}{@{\extracolsep{\fill}}llrrrl@{}}
\toprule
\textbf{Tier} & \textbf{Source} & \textbf{Timesteps} & \textbf{Episodes} & \textbf{Avg.\ Ep.\ Len.} & \textbf{Origin} \\
\midrule
1 & Human (on robot)                &    80,591 &    111 & $\sim$727 & Lab tournament \\
2 & Augmented                       &   322,364 &    444 & $\sim$727 & Symmetry 4$\times$ \\
3 & Synthetic (HW-validated policy) & 5,000,000 & 34,649 & $\sim$144 & PPO self-play \\
\bottomrule
\end{tabular*}
\end{table}
\section{Diagnostic Baselines}
\label{sec:baselines}

The baselines below are not intended to solve Klask. They are intended to \emph{fail informatively}, each revealing a specific algorithmic limitation. Table~\ref{tab:baselines_summary} previews the three failure modes; the rest of this section provides the supporting evidence.

\textbf{Action bounds.} The environment imposes no action limit, so each method sets
its own. The PPO expert that generates Tier~3 is clipped at $\pm0.2$\,m/s, with 16.6\%
of its commands exactly on that bound. BC is effectively unclipped: 4.1\% of its
commands exceed $0.2$\,m/s, with a maximum of $0.331$\,m/s. GAIL and BC+RL optimize at
$\pm1.0$\,m/s, and GAIL saturates at exactly $\pm1.0$ on 99.65\% of steps. The actuator
model was identified on commands within $\pm0.2$\,m/s~\citep{VenkatapathyA-RSS-26} \footnote{The deployed checkpoint's identification data spans commanded velocities within $\pm0.2$\,m/s; the wider range quoted in [7] refers to an earlier identification round.}, so
only BC operates near that range; GAIL and BC+RL command velocities up to five times
beyond it, which Table~\ref{tab:baselines_summary} reflects.

\begin{table}[h]
\caption{Diagnostic signature of each baseline. HW-compat. indicates that a method's commands stay within the $\pm 0.2$\,m/s range over which the actuator model was identified~\citep{VenkatapathyA-RSS-26}, so an improved baseline would be deployable without modification. GAIL and BC+RL optimize at $\pm 1.0$\,m/s, five times that range; their state/action interface matches, but their commands extrapolate beyond the identification data. $^{*}$4.1\% of BC's commands marginally exceed the range (max 0.331\,m/s). BC and GAIL metrics are self-play; BC+RL metrics are against the frozen PPO expert over five seeds.}
\label{tab:baselines_summary}
\centering
\small
\begin{tabularx}{\textwidth}{@{}lcccL@{}}
\toprule
\textbf{Method} & \textbf{Goals/Ep.} & \textbf{Player-in-Goal} & \textbf{HW-compat.} & \textbf{Diagnostic Failure Mode} \\
\midrule
BC (human, 80k)  & 0.102 & 0.157 & \checkmark$^{*}$ & Data-scaling plateau: covariate-shift floor \\
GAIL             & 0.116 & 0.201 & $\times$ & Interpretable but flat discriminator \\
BC+RL (sparse)   & 0.138 & 0.130 & $\times$ & Asymmetric prior: stabilizes sparse RL \\
BC+RL (dense)    & 0.128 & 0.067 & $\times$ & Asymmetric prior: no detectable effect under dense shaping \\
\bottomrule
\end{tabularx}
\end{table}

\subsection{Expert Policy Training}
\label{sec:expert_training}
The synthetic expert is trained with PPO using time-decaying reward shaping and enhanced self-play~\citep{openai2019dota2largescale}:
$r_{\text{total}}(t) = r_{\text{term}} + \sum_i R_i \cdot \exp(-t/N) \cdot r_{\text{shape},i}$.
Termination rewards are $\pm 1$ for goals; shaping includes ball-to-goal distance, peg-to-ball proximity, and ball speed, all decayed over training. Opponents are sampled from a checkpoint pool with geometric decay favoring recent policies. Full hyperparameters in Appendix~\ref{app:hyperparams}. The policy follows the training curriculum of~\citep{VenkatapathyA-RSS-26}; the checkpoint used here is deployed zero-shot on hardware in Section~\ref{sec:hardware}.

\subsection{Behavioral Cloning: Data-Scaling Plateau}
\label{sec:bc_scaling}

We train a two-layer MLP (64 units per layer) with entropy-regularized NLL: $\mathcal{L}_{\text{BC}} = -\log \pi_\theta(a \mid s) - \lambda_{\text{ent}} H(\pi_\theta) + \lambda_{L_2} \|\theta\|_2^2$, with hyperparameters tuned via Optuna (lr $6.72{\times}10^{-4}$, batch 128, 20 epochs).

\begin{figure}[h]
\centering
\includegraphics[width=0.85\linewidth,height=4.8cm,keepaspectratio]{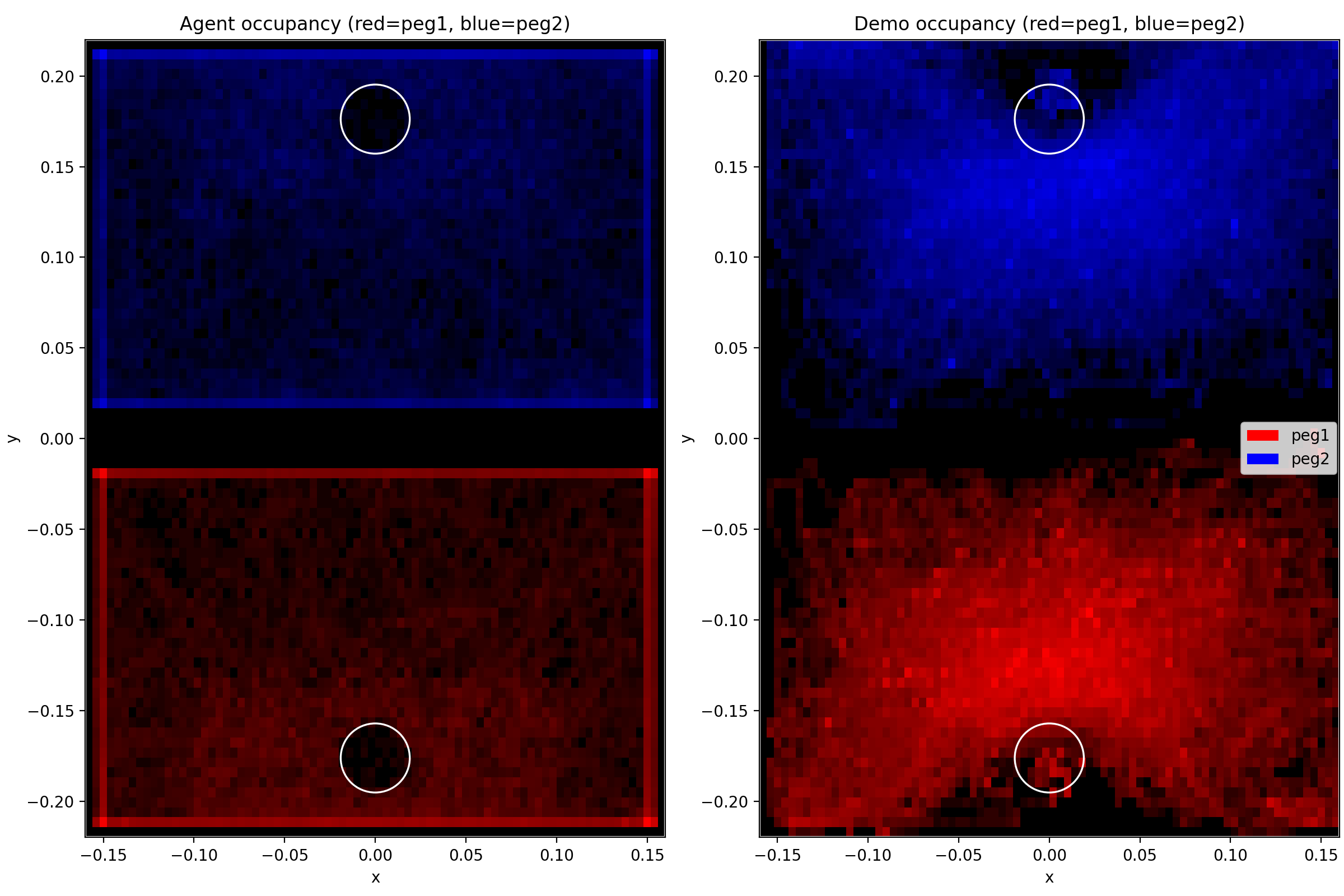}
    \caption{Peg occupancy: BC rollout (left) vs.\ human demonstrations (right). The BC policy occupies wall regions; expert demonstrations do not.}
    \label{fig:bc_heatmap}
\end{figure}

Trained on 80k human timesteps, the policy learns to track the ball and react to direction changes, but occupancy analysis (Figure~\ref{fig:bc_heatmap}) tells a different story: peg positions collapse to the board boundaries. This is textbook covariate shift: once a collision pushes the peg into an unvisited region, the policy has no recovery strategy and drifts to the wall.

\begin{figure}[h]
\centering
\includegraphics[width=0.8\linewidth,height=5cm,keepaspectratio]{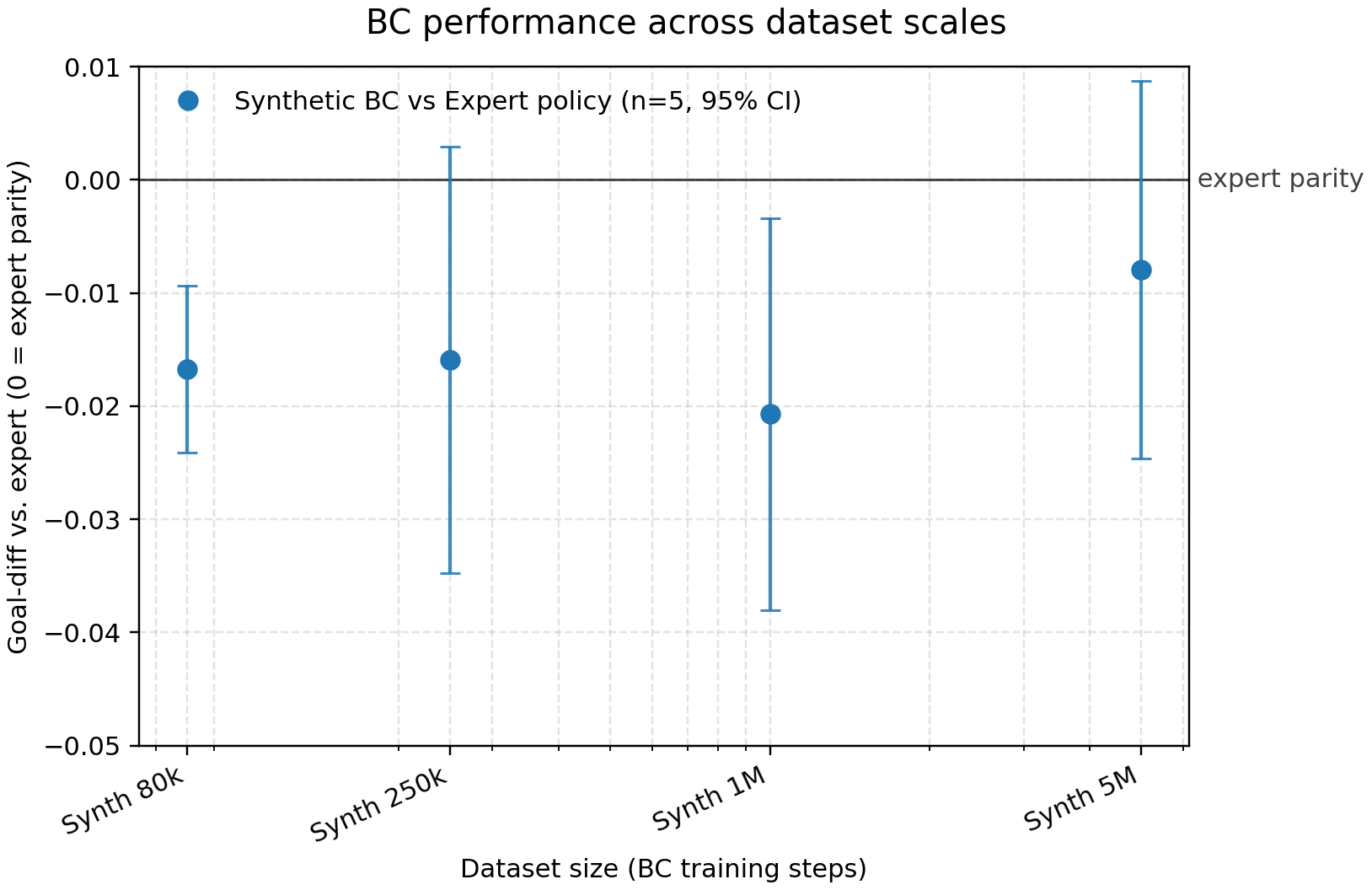}
\caption{BC data scaling on KlaskTron-IL. Performance sits roughly $0.015$ goal differential per episode below the expert across all four scales, with overlapping $95\,\%$ CIs ($n = 5$ seeds per scale). $60\,\times$ more demonstration data does not close the gap.}
\label{fig:scaling}
\end{figure}

The more revealing result comes from the scaling study. We train BC on each tier of synthetic data (80k through 5M), with 5 seeds per scale. As shown in Figure~\ref{fig:scaling}, mean performance is statistically flat across the full $60 \times$ range: all four scales sit roughly $0.015$ goal differential per episode below the expert, with overlapping $95\,\%$ confidence intervals.
The 5M agent is a noticeably better imitator – more ball contact, more reactive movement – but does not close the task-level gap. Held-out predictive $R^2$ on expert states rises monotonically with demonstrations (0.44/0.57/0.62/0.67 at 80k/250k/1M/5M, disjoint 95\% CIs) while this closed-loop gap stays flat, a pattern consistent with compounding error. Isolating the opponent's contribution remains future work. An algorithm that addresses covariate shift at its root should break this plateau and close the gap to the expert parity; we expect that DAgger variants, distributionally-robust BC~\citep{NEURIPS2024_c556da88}, or diffusion policies~\citep{Chi-RSS-23} are needed to break the pattern.

\subsection{GAIL: Discriminator as Feature Discovery}
\label{sec:gail}

We train GAIL~\citep{GAIL} with a \texttt{BasicShapedRewardNet} using potential-based shaping, a PPO generator, and several stabilization measures: freezing the discriminator when its accuracy on generated samples exceeds 0.75, reduced parallelism, and event-aware weighting that gives goal transitions higher importance. Training runs for $\sim$5M steps (lr $2.15{\times}10^{-4}$, batch 2048, $\gamma{=}0.99$).

Training is numerically stable (discriminator loss $\approx$0.68, accuracy 55--60\%), but the resulting gameplay is poor. The agent collapses to passive center-seeking behavior and achieves a goals ratio of only 1.14 in self-play (Table~\ref{tab:eval}), marginally above BC's 1.05 but with more peg-in-goal terminations.

\begin{figure}[h]
\centering
\includegraphics[width=0.75\linewidth,height=5.5cm,keepaspectratio]{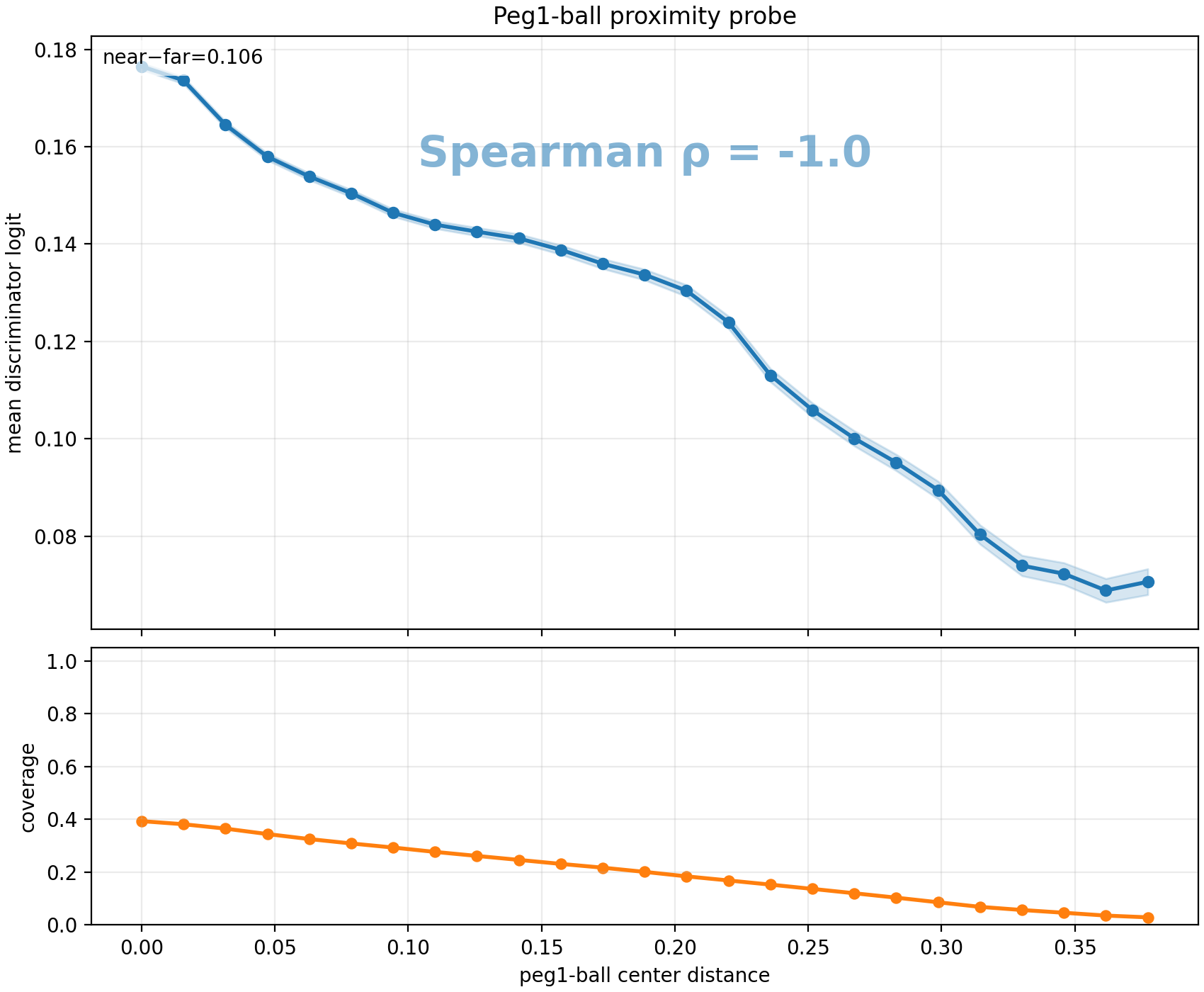}
\caption{Discriminator probing. Top: logit vs.\ peg--ball distance (Spearman $\rho{=}{-}1.0$). Bottom: data coverage. The perfect rank ordering confirms feature discovery; the narrow logit range is consistent with the weak reward signal under this configuration.}
\label{fig:disc_sensitivity}
\end{figure}

The informative result comes from probing the discriminator directly. We take expert validation transitions, fix all state variables, and sweep only peg--ball distance, re-scoring each configuration. Figure~\ref{fig:disc_sensitivity} shows a monotonic decrease in the raw logit as distance increases (Spearman $\rho = -1.0$): the discriminator has learned that staying close to the ball is an expert-like feature. The problem is that the logit span is only 0.00 to 0.09, which maps to sigmoid outputs between 0.50 and 0.52. PPO did not exploit this signal in our configuration. Whether the narrow span is the cause is confounded: expert and generator action scales differ markedly ($\pm0.2$ vs.\ $\pm1.0$\,m/s, Section~\ref{sec:baselines}), so the discriminator may separate them on action magnitude alone. We therefore read the sweep as a state-sensitivity analysis, not a causal account of rollout failure or a fundamental GAIL limitation.

\begin{table}[h]
\caption{Policy evaluation. Mean $\pm$ sample SD across 3 random seeds  of a single trained policy per method (1{,}000 episodes per repeat).}
\label{tab:eval}
\centering
\small
\begin{tabular}{@{}lccc@{}}
\toprule
Metric & BC$\leftrightarrow$BC & GAIL$\leftrightarrow$GAIL & GAIL$\leftrightarrow$BC \\
\midrule
Goals ratio    & 1.053 $\pm$ 0.278 & 1.137 $\pm$ 0.043 & 1.357 $\pm$ 0.154 \\
Goals scored/ep & 0.102 $\pm$ 0.013 & 0.116 $\pm$ 0.008 & 0.144 $\pm$ 0.007 \\
Timeout rate   & 0.472 $\pm$ 0.024 & 0.309 $\pm$ 0.005 & 0.379 $\pm$ 0.023 \\
Player-in-Goal & 0.157 $\pm$ 0.006 & 0.201 $\pm$ 0.007 & 0.197 $\pm$ 0.016 \\
Ep.\ length    & 146.8 $\pm$ 4.6 & 103.6 $\pm$ 1.7 & 128.0 $\pm$ 6.8 \\
\bottomrule
\end{tabular}
\end{table}

The discriminator operates on single-frame state--action pairs, limiting it to static spatial features; equipping it with observation history is a natural next step, and more expressive discriminators such as diffusion-based ones~\citep{DRAIL} target its brittleness. Practically, the peg--ball proximity feature the discriminator identifies with perfect rank correlation directly validates the \texttt{distance\_player\_ball} shaping term used in the expert policy's reward. When reward engineering is the bottleneck, probing a GAIL discriminator can substitute for hand-tuning shaping terms even when the discriminator itself fails as a reward source.

\subsection{BC-Bootstrapped RL: Asymmetric Initialization}
\label{sec:ablation}

We run a $2{\times}2$ factorial experiment: \{random, BC-pretrained\} initialization crossed with \{sparse ($\pm 10$ for goals), dense (multi-term shaping)\} reward. All conditions use PPO and are evaluated against the frozen PPO expert over 1000 episodes per seed. Confidence intervals are taken over training seeds.

\begin{figure}[h]
\centering
\includegraphics[width=0.95\linewidth,height=5cm,keepaspectratio]{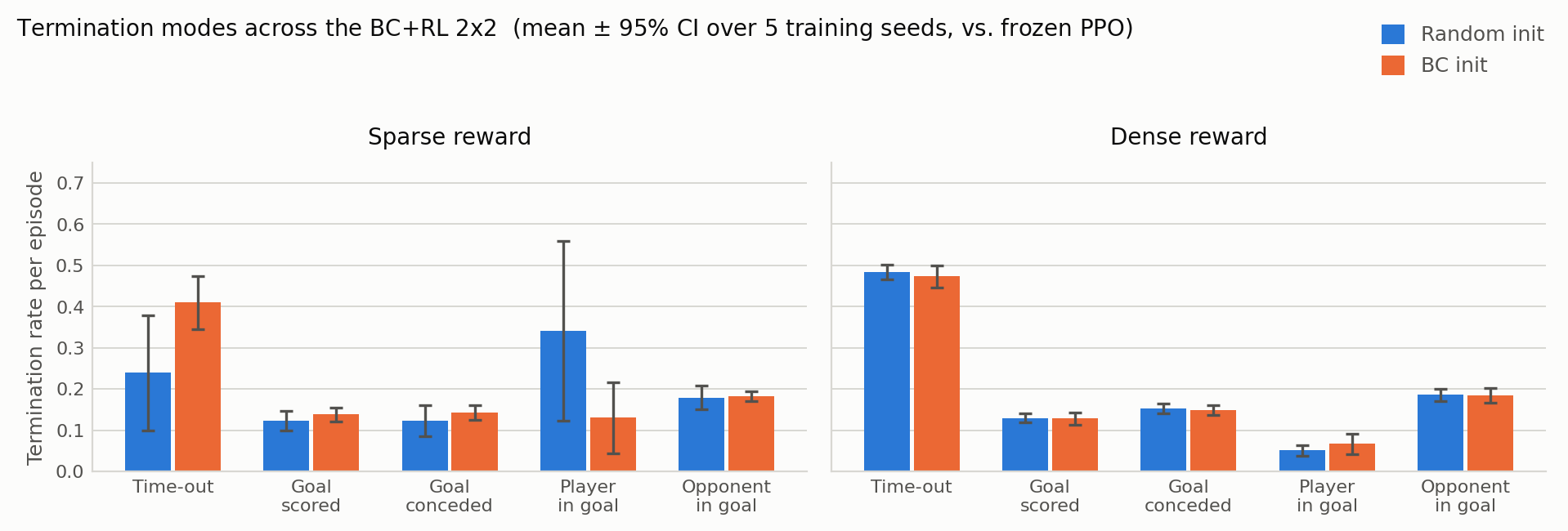}
\caption{Termination modes across the $2{\times}2$ ablation, five training seeds per condition with 95\% CIs, evaluated against the frozen PPO expert. BC initialization is a strong stabilizer under sparse rewards. Under dense reward the two initializations are indistinguishable.}
\label{fig:ablation}
\end{figure}

Under \textbf{sparse rewards}, BC initialization cuts peg-in-goal terminations from 34\% to 13\% (a 62\% relative reduction), extends average episode length from 97.9 to 136.3 steps ($+38.4$, 95\% CI $[+2.2, +74.5]$), and lifts goal scoring from 0.123 to 0.138 per episode (Figure~\ref{fig:ablation}). The BC prior prevents the agent from wandering into its own goal during early training, giving RL a stable behavioral base. The benefit is uneven across seeds: two of five randomly initialized runs collapse into self-scoring (0.564, 0.485) while three do not (0.176--0.290), whereas every BC-initialized seed stays no higher than 0.240. Because of that spread, only the episode-length contrast clears 95\% at $n=5$.

Under \textbf{dense rewards}, we detect no effect. Peg-in-goal ($0.051 \to 0.067$), episode length ($153.2 \to 149.9$), and goal scoring ($0.130 \to 0.128$) all move within noise, and every contrast's 95\% interval contains zero. A preliminary single-seed comparison against a fixed BC opponent suggested that the BC prior degraded dense-reward performance; that effect survives neither the change of evaluation opponent nor the addition of seeds.

The prior therefore helps where the reward alone cannot: it suppresses early self-scoring when goals are the only signal, and adds nothing once dense shaping already supplies that gradient. The open question is consistency rather than average gain — two of five random seeds collapsed and three did not, so the useful advance would be making the sparse-reward benefit reliable across seeds. Methods that scale the demonstration prior with reward-signal strength, such as KL-regularized fine-tuning or advantage-weighted regression, are the natural candidates.

\subsection{Zero-Shot Hardware Deployment of a BC Baseline}
\label{sec:hardware}

The hardware path described in Section~\ref{sec:platform} was validated for the PPO expert~\citep{VenkatapathyA-RSS-26}. To test whether an imitation-learned policy inherits it, we deployed the BC policy trained on the full 5M-step Tier~3 prefix (BC-5M, Section~\ref{sec:bc_scaling}) on the physical robot zero-shot, without fine-tuning. The interface is unchanged: the 12D EKF state as observation, a 2D velocity command as action, and the same $\pm0.2$\,m/s per-axis command limit the PPO expert is trained with.

\textbf{Protocol.} We ran two matched sessions against the same human opponent, one with BC-5M and one with the PPO expert, totalling 7.3\,min of play (21,818 steps at 50\,Hz). For each policy we compare its hardware rollouts with its simulated rollouts on eight derived geometric features and on the commanded speed, using the 1-Wasserstein distance between the two distributions normalised by each feature's 1--99\% range, so that both policies are measured identically.

\textbf{Results.} BC-5M shifts less from simulation to hardware than the PPO expert (normalised $W_1$ of 0.05 vs.\ 0.15). Two behavioural differences from the expert are visible on hardware, both with disjoint 95\% confidence intervals: BC-5M engages the ball 31\% less often, and after losing contact it re-engages more slowly (90th percentile 5.6\,s vs.\ 3.4\,s). Both are consistent with the limited recovery behaviour observed in simulation (Section~\ref{sec:bc_scaling}), so for this policy the simulated diagnostic carries over to the robot.

\textbf{Caveat.} For both policies, realised peg speed on hardware is roughly half the simulated value at equal command. This is a property of the inherited actuator model rather than of the IL policy; it affects PPO and BC alike and leaves the comparison between them intact, but absolute simulated speeds should not be read literally on hardware.

\textbf{Scope.} Zero-shot transfer is now demonstrated for the PPO expert and for this one BC policy, not for IL methods in general. GAIL and BC+RL have not been deployed; their $\pm1.0$\,m/s commands are five times the limit under which transfer has been demonstrated (Table~\ref{tab:baselines_summary}).

\section{Limitations}
\label{sec:limitations}

\textbf{Observation space.} The benchmark exposes a 12D state vector (positions and velocities of both pegs and the ball). Methods designed around visuomotor inputs will need either a learned perception front-end or a different observation interface; the simulator can render images, but no curated image dataset is part of the current release.

\textbf{Rule simplification.} Magnetic ``biscuit'' obstacles are omitted to isolate contact-rich and adversarial dynamics from combinatorial rule complexity. The benchmark does not exercise the full strategic depth of the underlying game.

\textbf{Tier 1 action reconstruction.} Human action labels are derived from finite-differenced position estimates, which makes them nearly deterministically predictable from consecutive observations. The mitigation -- zeroing velocity components of the state during BC training -- is documented in Section~\ref{sec:dataset}; users should apply it or verify their method is invariant to this leakage channel.

\textbf{Synthetic expert quality.} The PPO expert reaches an Elo of 1283 (61.8\% win rate vs.\ a 1200 baseline, 10\% vs.\ a recreational human). It is a strong-but-imperfect expert and does not represent ceiling performance.

\textbf{Evaluation uncertainty.} The BC and GAIL entries in Table~\ref{tab:eval} are three evaluation repeats of a single trained policy each, so their error bars reflect rollout variance rather than training variance. The BC scaling study uses five training seeds per scale with 95\% CIs; the plateau is robust to initialization but has not been replicated across different BC architectures or hyperparameter settings.

\textbf{Hardware coverage of the IL baselines.} Zero-shot transfer has been demonstrated for the PPO expert~\citep{VenkatapathyA-RSS-26} and for a single BC policy (Section~\ref{sec:hardware}), not for IL methods in general. GAIL and BC+RL have not been deployed, and their $\pm1.0$\,m/s commands are five times the limit under which transfer has been demonstrated (Table~\ref{tab:baselines_summary}). Realised peg speed on hardware is about half the simulated value at equal command for both deployed policies, so simulated speeds should not be read literally on the robot.

\section{Conclusion}
\label{sec:conclusion}

KlaskTron-IL contributes the imitation-learning layer to a physical adversarial multi-agent platform that already has a validated zero-shot sim-to-real bridge. The three failure modes we expose -- BC's data-scaling plateau, GAIL's interpretable but weak-gradient discriminator, and the asymmetric BC-prior $\times$ reward-density interaction -- are diagnostic in the strict sense: they isolate specific algorithmic limitations rather than generic difficulty. Because the policy interface throughout the benchmark is identical to the one the actuator network bridges to hardware, an algorithm that breaks any of these failure modes while keeping its commands within the actuator model's identified action range is a candidate for deployment on the physical robot. Section~\ref{sec:hardware} demonstrates this path zero-shot for the strongest BC baseline. We see this as the central feature differentiating KlaskTron-IL from cooperative IL benchmarks: progress here translates to a real-world artifact rather than to better simulator numbers. Future work will reintroduce the biscuit obstacles, equip the discriminator with observation history, and -- most importantly -- close the loop by deploying improved IL baselines on the physical platform.

\clearpage
\acknowledgments{The first author thanks Aswin Ramachandran for his supervision and guidance throughout this project, and the KlaskTron platform team -- Jona Schulz, Maurus Derungs, Carlo Angelini and Tobias Meier -- whose hardware, simulator and actuator model this work builds on. We are grateful to colleagues and students at IDSC for discussions and advice and to everyone who took part in the lab tournament that produced the Tier 1 human demonstrations.}

\newpage
\appendix

\section{Training Hyperparameters}
\label{app:hyperparams}
The expert's PPO hyperparameters are those of~\citep{VenkatapathyA-RSS-26}; the exact rl\_games configuration (\texttt{agent.yaml}) ships with the code release alongside the checkpoint.

\textbf{Self-play.} Opponent sampling uses geometric decay: $w_i = 0.9^{(N - i - 1)}$, where $i$ indexes checkpoints (most recent $= N{-}1$). The pool updates when the average score difference exceeds 0.7.

\section{Reward Configuration}
\label{app:reward}

Table~\ref{tab:reward_terms} lists the active reward terms used in the dense RL condition. The sparse condition retains only \texttt{goal\_scored} and \texttt{goal\_conceded} with weights $\pm 10$.

\begin{table}[h]
\caption{Dense reward terms and weights used in the BC+RL ablation.}
\label{tab:reward_terms}
\centering
\small
\begin{tabular}{@{}lr@{}}
\toprule
\textbf{Reward term} & \textbf{Weight} \\
\midrule
\texttt{goal\_scored}                  & $+10.0$ \\
\texttt{goal\_conceded}                & $-10.0$ \\
\texttt{player\_in\_goal}              & $-10.0$ \\
\texttt{opponent\_in\_goal}            & $-10.0$ \\
\texttt{distance\_ball\_opponent\_goal} & $+1.0$  \\
\texttt{collision\_player\_ball}        & $+0.5$  \\
\texttt{shot\_over\_middle\_line}       & $+0.1$  \\
\texttt{ball\_speed}                    & $+0.1$  \\
\bottomrule
\end{tabular}
\end{table}


\begin{thebibliography}{24}
\providecommand{\natexlab}[1]{#1}
\providecommand{\url}[1]{\texttt{#1}}
\expandafter\ifx\csname urlstyle\endcsname\relax
  \providecommand{\doi}[1]{doi: #1}\else
  \providecommand{\doi}{doi: \begingroup \urlstyle{rm}\Url}\fi

\bibitem[D’Ambrosio et~al.(2025)D’Ambrosio, Abeyruwan, Graesser, Iscen,
  {Ben Amor}, Bewley, Reed, Reymann, Takayama, Tassa, Choromanski, Coumans,
  Jain, Jaitly, Jaques, Kataoka, Kuang, Lazic, Mahjourian, Moore, Oslund,
  Shankar, Sindhwani, Vanhoucke, Vesom, Xu, and Sanketi]{table_tennis}
D.~B. D’Ambrosio, S.~Abeyruwan, L.~Graesser, A.~Iscen, H.~{Ben Amor},
  A.~Bewley, B.~J. Reed, K.~Reymann, L.~Takayama, Y.~Tassa, K.~Choromanski,
  E.~Coumans, D.~Jain, N.~Jaitly, N.~Jaques, S.~Kataoka, Y.~Kuang, N.~Lazic,
  R.~Mahjourian, S.~Moore, K.~Oslund, A.~Shankar, V.~Sindhwani, V.~Vanhoucke,
  G.~Vesom, P.~Xu, and P.~R. Sanketi.
\newblock Achieving human level competitive robot table tennis.
\newblock In \emph{2025 IEEE International Conference on Robotics and
  Automation (ICRA)}, pages 74--82, 2025.
\newblock \doi{10.1109/ICRA55743.2025.11127501}.

\bibitem[Mandlekar et~al.(2022)Mandlekar, Xu, Wong, Nasiriany, Wang, Kulkarni,
  Fei-Fei, Savarese, Zhu, and Mart\'in-Mart\'in]{pmlr-v164-mandlekar22a}
A.~Mandlekar, D.~Xu, J.~Wong, S.~Nasiriany, C.~Wang, R.~Kulkarni, L.~Fei-Fei,
  S.~Savarese, Y.~Zhu, and R.~Mart\'in-Mart\'in.
\newblock What matters in learning from offline human demonstrations for robot
  manipulation.
\newblock In A.~Faust, D.~Hsu, and G.~Neumann, editors, \emph{Proceedings of
  the 5th Conference on Robot Learning}, volume 164 of \emph{Proceedings of
  Machine Learning Research}, pages 1678--1690. PMLR, 08--11 Nov 2022.
\newblock URL \url{https://proceedings.mlr.press/v164/mandlekar22a.html}.

\bibitem[Walke et~al.(2023)Walke, Black, Zhao, Vuong, Zheng, Hansen-Estruch,
  He, Myers, Kim, Du, Lee, Fang, Finn, and Levine]{pmlr-v229-walke23a}
H.~R. Walke, K.~Black, T.~Z. Zhao, Q.~Vuong, C.~Zheng, P.~Hansen-Estruch, A.~W.
  He, V.~Myers, M.~J. Kim, M.~Du, A.~Lee, K.~Fang, C.~Finn, and S.~Levine.
\newblock Bridgedata v2: A dataset for robot learning at scale.
\newblock In J.~Tan, M.~Toussaint, and K.~Darvish, editors, \emph{Proceedings
  of The 7th Conference on Robot Learning}, volume 229 of \emph{Proceedings of
  Machine Learning Research}, pages 1723--1736. PMLR, 06--09 Nov 2023.
\newblock URL \url{https://proceedings.mlr.press/v229/walke23a.html}.

\bibitem[Jia et~al.(2024)Jia, Blessing, Jiang, Reuss, Donat, Lioutikov, and
  Neumann]{jia2024towards}
X.~Jia, D.~Blessing, X.~Jiang, M.~Reuss, A.~Donat, R.~Lioutikov, and
  G.~Neumann.
\newblock Towards diverse behaviors: A benchmark for imitation learning with
  human demonstrations.
\newblock In \emph{The Twelfth International Conference on Learning
  Representations}, 2024.
\newblock URL \url{https://openreview.net/forum?id=6pPYRXKPpw}.

\bibitem[{KLASK}(2024)]{Klask}
{KLASK}.
\newblock Official web page of klask.
\newblock \url{https://www.klaskgame.com/pages/about-us}, 2024.

\bibitem[Hwangbo et~al.(2019)Hwangbo, Lee, Dosovitskiy, Bellicoso, Tsounis,
  Koltun, and Hutter]{hwangbo}
J.~Hwangbo, J.~Lee, A.~Dosovitskiy, D.~Bellicoso, V.~Tsounis, V.~Koltun, and
  M.~Hutter.
\newblock Learning agile and dynamic motor skills for legged robots.
\newblock \emph{Science Robotics}, 4\penalty0 (26):\penalty0 eaau5872, 2019.
\newblock \doi{10.1126/scirobotics.aau5872}.
\newblock URL
  \url{https://www.science.org/doi/abs/10.1126/scirobotics.aau5872}.

\bibitem[Venkatapathy et~al.(2026)Venkatapathy, Schulz, Derungs, Angelini,
  Meier, and D'Andrea]{VenkatapathyA-RSS-26}
A.~K.~R. Venkatapathy, J.~Schulz, M.~Derungs, C.~Angelini, T.~Meier, and
  R.~D'Andrea.
\newblock {KlaskTron: An Open-Source Platform for Physical Adversarial
  Multi-Agent RL}.
\newblock In \emph{Proceedings of Robotics: Science and Systems}, Sydney,
  Australia, July 2026.
\newblock \doi{10.15607/RSS.2026.XXII.038}.

\bibitem[Gavenski et~al.(2024)Gavenski, Luck, and
  Rodrigues]{gavenski2024ildatasets}
N.~Gavenski, M.~Luck, and O.~Rodrigues.
\newblock Imitation learning datasets: A toolkit for creating datasets,
  training agents and benchmarking.
\newblock In \emph{Proceedings of the 23rd International Conference on
  Autonomous Agents and Multiagent Systems}, AAMAS '24, pages 2800--2802,
  Richland, SC, 2024. International Foundation for Autonomous Agents and
  Multiagent Systems.
\newblock ISBN 9798400704864.

\bibitem[Fu et~al.(2021)Fu, Kumar, Nachum, Tucker, and
  Levine]{fu2021d4rldatasetsdeepdatadriven}
J.~Fu, A.~Kumar, O.~Nachum, G.~Tucker, and S.~Levine.
\newblock D4rl: Datasets for deep data-driven reinforcement learning, 2021.
\newblock URL \url{https://arxiv.org/abs/2004.07219}.

\bibitem[Gulcehre et~al.(2020)Gulcehre, Wang, Novikov, Paine, G\'{o}mez, Zolna,
  Agarwal, Merel, Mankowitz, Paduraru, Dulac-Arnold, Li, Norouzi, Hoffman,
  Heess, and de~Freitas]{RL_Unplugged}
C.~Gulcehre, Z.~Wang, A.~Novikov, T.~Paine, S.~G\'{o}mez, K.~Zolna, R.~Agarwal,
  J.~S. Merel, D.~J. Mankowitz, C.~Paduraru, G.~Dulac-Arnold, J.~Li,
  M.~Norouzi, M.~Hoffman, N.~Heess, and N.~de~Freitas.
\newblock Rl unplugged: A suite of benchmarks for offline reinforcement
  learning.
\newblock In H.~Larochelle, M.~Ranzato, R.~Hadsell, M.~Balcan, and H.~Lin,
  editors, \emph{Advances in Neural Information Processing Systems}, volume~33,
  pages 7248--7259. Curran Associates, Inc., 2020.
\newblock URL
  \url{https://proceedings.neurips.cc/paper_files/paper/2020/file/51200d29d1fc15f5a71c1dab4bb54f7c-Paper.pdf}.

\bibitem[Younis et~al.(2024)Younis, Perez-Vicente, Balis, Dudley, Davey, and
  Terry]{younis_2024_13767625}
O.~G. Younis, R.~Perez-Vicente, J.~U. Balis, W.~Dudley, A.~Davey, and J.~K.
  Terry.
\newblock Minari, Sept. 2024.
\newblock URL \url{https://doi.org/10.5281/zenodo.13767625}.

\bibitem[Bi and D’Andrea(2024)]{sample_efficient}
T.~Bi and R.~D’Andrea.
\newblock Sample-efficient learning to solve a real-world labyrinth game using
  data-augmented model-based reinforcement learning.
\newblock In \emph{2024 IEEE International Conference on Robotics and
  Automation (ICRA)}, pages 7455--7460, 2024.
\newblock \doi{10.1109/ICRA57147.2024.10610577}.

\bibitem[Jankowski et~al.(2024)Jankowski, Marić, and
  Calinon]{jankowski2024airlihockeyhighlyreactivecontact}
J.~Jankowski, A.~Marić, and S.~Calinon.
\newblock Airlihockey: Highly reactive contact control and stochastic optimal
  shooting, 2024.
\newblock URL \url{https://arxiv.org/abs/2401.14964}.

\bibitem[Moos et~al.(2024)Moos, Derstroff, Schröder, and Clever]{Moos_2024}
J.~Moos, C.~Derstroff, N.~Schröder, and D.~Clever.
\newblock Learning to play foosball: System and baselines.
\newblock In \emph{2024 IEEE International Conference on Robotics and
  Automation (ICRA)}, pages 4326--4332, 2024.
\newblock \doi{10.1109/ICRA57147.2024.10611321}.

\bibitem[Ross et~al.(2011)Ross, Gordon, and Bagnell]{Dagger}
S.~Ross, G.~Gordon, and D.~Bagnell.
\newblock A reduction of imitation learning and structured prediction to
  no-regret online learning.
\newblock In G.~Gordon, D.~Dunson, and M.~Dudík, editors, \emph{Proceedings of
  the Fourteenth International Conference on Artificial Intelligence and
  Statistics}, volume~15 of \emph{Proceedings of Machine Learning Research},
  pages 627--635, Fort Lauderdale, FL, USA, 11--13 Apr 2011. PMLR.
\newblock URL \url{https://proceedings.mlr.press/v15/ross11a.html}.

\bibitem[Hoque et~al.(2022)Hoque, Balakrishna, Novoseller, Wilcox, Brown, and
  Goldberg]{thriftydagger}
R.~Hoque, A.~Balakrishna, E.~Novoseller, A.~Wilcox, D.~S. Brown, and
  K.~Goldberg.
\newblock Thriftydagger: Budget-aware novelty and risk gating for interactive
  imitation learning.
\newblock In A.~Faust, D.~Hsu, and G.~Neumann, editors, \emph{Proceedings of
  the 5th Conference on Robot Learning}, volume 164 of \emph{Proceedings of
  Machine Learning Research}, pages 598--608. PMLR, 08--11 Nov 2022.
\newblock URL \url{https://proceedings.mlr.press/v164/hoque22a.html}.

\bibitem[Hoque et~al.(2021)Hoque, Balakrishna, Putterman, Luo, Brown, Seita,
  Thananjeyan, Novoseller, and Goldberg]{lazydagger}
R.~Hoque, A.~Balakrishna, C.~Putterman, M.~Luo, D.~S. Brown, D.~Seita,
  B.~Thananjeyan, E.~Novoseller, and K.~Goldberg.
\newblock Lazydagger: Reducing context switching in interactive imitation
  learning.
\newblock In \emph{2021 IEEE 17th International Conference on Automation
  Science and Engineering (CASE)}, pages 502--509, 2021.
\newblock \doi{10.1109/CASE49439.2021.9551469}.

\bibitem[Mehta et~al.(2025)Mehta, Ciftci, Ramachandran, Bansal, and
  Losey]{Stable_BC}
S.~A. Mehta, Y.~U. Ciftci, B.~Ramachandran, S.~Bansal, and D.~P. Losey.
\newblock Stable-bc: Controlling covariate shift with stable behavior cloning.
\newblock \emph{IEEE Robotics and Automation Letters}, 10\penalty0
  (2):\penalty0 1952--1959, 2025.
\newblock \doi{10.1109/LRA.2025.3526439}.

\bibitem[Seo et~al.(2024)Seo, Lee, Lee, Hwang, Yang, and
  Kim]{NEURIPS2024_c556da88}
S.~Seo, B.-J. Lee, J.~Lee, H.~Hwang, H.~Yang, and K.-E. Kim.
\newblock Mitigating covariate shift in behavioral cloning via robust
  stationary distribution correction.
\newblock In A.~Globerson, L.~Mackey, D.~Belgrave, A.~Fan, U.~Paquet,
  J.~Tomczak, and C.~Zhang, editors, \emph{Advances in Neural Information
  Processing Systems}, volume~37, pages 109177--109201. Curran Associates,
  Inc., 2024.
\newblock \doi{10.52202/079017-3466}.
\newblock URL
  \url{https://proceedings.neurips.cc/paper_files/paper/2024/file/c556da88a2665e6266453d8c9b8a552d-Paper-Conference.pdf}.

\bibitem[Chi et~al.(2023)Chi, Feng, Du, Xu, Cousineau, Burchfiel, and
  Song]{Chi-RSS-23}
C.~Chi, S.~Feng, Y.~Du, Z.~Xu, E.~Cousineau, B.~C. Burchfiel, and S.~Song.
\newblock {Diffusion Policy: Visuomotor Policy Learning via Action Diffusion}.
\newblock In \emph{Proceedings of Robotics: Science and Systems}, Daegu,
  Republic of Korea, July 2023.
\newblock \doi{10.15607/RSS.2023.XIX.026}.

\bibitem[Ho and Ermon(2016)]{GAIL}
J.~Ho and S.~Ermon.
\newblock Generative adversarial imitation learning.
\newblock In D.~Lee, M.~Sugiyama, U.~Luxburg, I.~Guyon, and R.~Garnett,
  editors, \emph{Advances in Neural Information Processing Systems}, volume~29.
  Curran Associates, Inc., 2016.
\newblock URL
  \url{https://proceedings.neurips.cc/paper_files/paper/2016/file/cc7e2b878868cbae992d1fb743995d8f-Paper.pdf}.

\bibitem[Lai et~al.(2024)Lai, Wang, Hsieh, Wang, Chen, and Sun]{DRAIL}
C.-M. Lai, H.-C. Wang, P.-C. Hsieh, Y.-C.~F. Wang, M.-H. Chen, and S.-H. Sun.
\newblock Diffusion-reward adversarial imitation learning.
\newblock In A.~Globerson, L.~Mackey, D.~Belgrave, A.~Fan, U.~Paquet,
  J.~Tomczak, and C.~Zhang, editors, \emph{Advances in Neural Information
  Processing Systems}, volume~37, pages 95456--95487. Curran Associates, Inc.,
  2024.
\newblock \doi{10.52202/079017-3025}.
\newblock URL
  \url{https://proceedings.neurips.cc/paper_files/paper/2024/file/ad47b1801557e4be37d30baf623de426-Paper-Conference.pdf}.

\bibitem[Hansen et~al.(2023)Hansen, Lin, Su, Wang, Kumar, and
  Rajeswaran]{hansen2023modem}
N.~Hansen, Y.~Lin, H.~Su, X.~Wang, V.~Kumar, and A.~Rajeswaran.
\newblock Modem: Accelerating visual model-based reinforcement learning with
  demonstrations.
\newblock In \emph{The Eleventh International Conference on Learning
  Representations}, 2023.
\newblock URL \url{https://openreview.net/forum?id=JdTnc9gjVfJ}.

\bibitem[OpenAI et~al.(2019)OpenAI, Berner, Brockman, Chan, Cheung, Dębiak,
  Dennison, Farhi, Fischer, Hashme, Hesse, Józefowicz, Gray, Olsson, Pachocki,
  Petrov, d.~O.~Pinto, Raiman, Salimans, Schlatter, Schneider, Sidor,
  Sutskever, Tang, Wolski, and Zhang]{openai2019dota2largescale}
OpenAI, C.~Berner, G.~Brockman, B.~Chan, V.~Cheung, P.~Dębiak, C.~Dennison,
  D.~Farhi, Q.~Fischer, S.~Hashme, C.~Hesse, R.~Józefowicz, S.~Gray,
  C.~Olsson, J.~Pachocki, M.~Petrov, H.~P. d.~O.~Pinto, J.~Raiman, T.~Salimans,
  J.~Schlatter, J.~Schneider, S.~Sidor, I.~Sutskever, J.~Tang, F.~Wolski, and
  S.~Zhang.
\newblock Dota 2 with large scale deep reinforcement learning, 2019.
\newblock URL \url{https://arxiv.org/abs/1912.06680}.

\end{thebibliography}
\end{document}